\documentclass[runningheads]{llncs}

\usepackage{eccv}

\usepackage{eccvabbrv}

\usepackage{graphicx}
\usepackage{booktabs}

\usepackage[accsupp]{axessibility}  

\usepackage{hyperref}

\usepackage{orcidlink}

\begin{document}

\title{Exploring Conditional Multi-Modal Prompts for Zero-shot HOI Detection}


\author{Ting Lei\inst{1} \and
Shaofeng Yin\inst{1} \and
Yuxin Peng \orcidlink{0000-0001-7658-3845} \inst{1} \and
Yang Liu \orcidlink{0000-0002-4259-3882} \inst{1,2} \thanks{Corresponding author}}


\institute{Wangxuan Institute of Computer Technology, Peking University \and
State Key Laboratory of General Artificial Intelligence, Peking University \\
\email{\{ting\_lei,pengyuxin,yangliu\}@pku.edu.cn} \quad \email{yin\_shaofeng@stu.pku.edu.cn}}

\maketitle

\begin{abstract}
    Zero-shot Human-Object Interaction (HOI) detection has emerged as a frontier topic due to its capability to detect HOIs beyond a predefined set of categories. This task entails not only identifying the interactiveness of human-object pairs and localizing them but also recognizing both seen and unseen interaction categories.
    In this paper, we introduce a novel framework for zero-shot HOI detection using Conditional Multi-Modal Prompts, namely CMMP. This approach enhances the generalization of large foundation models, such as CLIP, when fine-tuned for HOI detection.
    Unlike traditional prompt-learning methods, we propose learning decoupled vision and language prompts for interactiveness-aware visual feature extraction and generalizable interaction classification, respectively. 
    Specifically, we integrate prior knowledge of different granularity into conditional vision prompts, including an input-conditioned instance prior and a global spatial pattern prior. The former encourages the image encoder to treat instances belonging to seen or potentially unseen HOI concepts equally while the latter provides representative plausible spatial configuration of the human and object under interaction. 
    Besides, we employ language-aware prompt learning with a consistency constraint to preserve the knowledge of the large foundation model to enable better generalization in the text branch.
    Extensive experiments demonstrate the efficacy of our detector with conditional multi-modal prompts, outperforming previous state-of-the-art on unseen classes of various zero-shot settings.
    The code and models are available at \url{https://github.com/ltttpku/CMMP}.
\end{abstract}

\section{Introduction}
\label{sec:intro}

Human-object interaction (HOI) detection has been introduced by~\cite{gupta2015visual} and plays an important role in understanding high-level human-centric scenes. Given an image, HOI detection aims to localize human and object pairs and recognize their interactions, \textit{i.e.} a set of $<$human, object, action$>$ triplets. Traditionally, human-object interaction detectors can be categorized as one- or two-stage. 
One-stage methods leverage multi-stream networks~\cite{liao2020ppdm,wang2020learning} or encoder-decoder architectures~\cite{zou2021end, chen2021reformulating, kim2021hotr, tamura2021qpic, zhong2022towards, liao2022gen, ning2023hoiclip} to predict HOI triplets from a holistic image context in an end-to-end manner. 
Two-stage methods~\cite{zhou2019relation, li2019transferable,liu2020amplifying,liu2020consnet,li2020hoi,zhang2021spatially,zhang2022efficient,zhang2022exploring, liu2022interactiveness,wu2022mining} first localize humans and objects separately using off-the-shelf detectors (\eg,\ DETR~\cite{carion2020end}), followed by utilizing the region features from the localized areas to predict the interaction class.

\begin{figure}[tb]
\centering
\subfloat[
HM: Harmonic Mean. The averaged performance on unseen classes and harmonic mean across all zero-shot settings of HICO-DET.
]{
\label{fig:degrade_uv}
\includegraphics[width=0.35\linewidth]{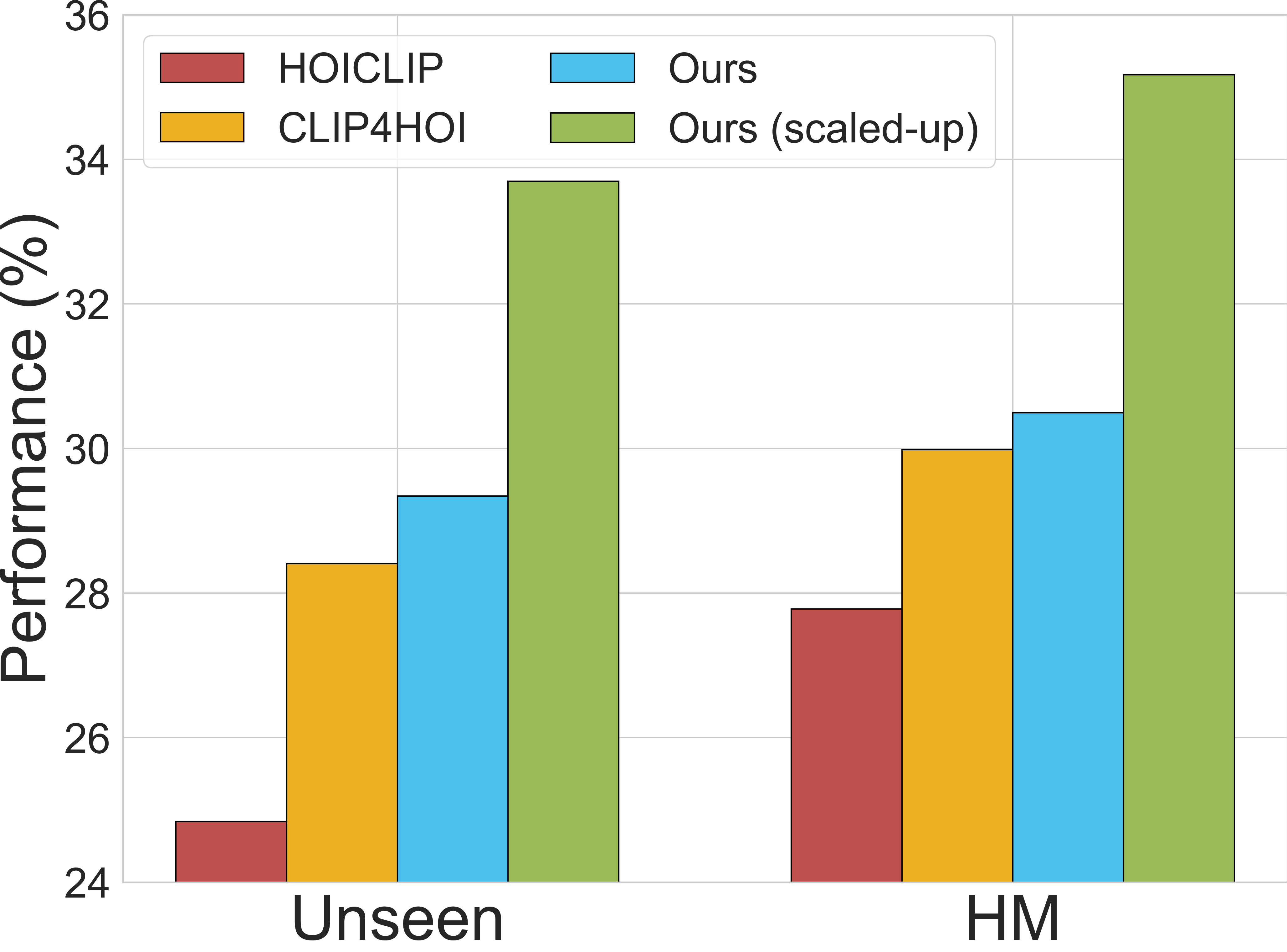}
} 
\hspace{1mm}
\subfloat[Left: Spatial cues help recognize the interactiveness of unseen HOI concepts; 
Right: The training trajectory of prompts.
]{
    \label{fig:teasor}
    \includegraphics[width=0.3\linewidth]{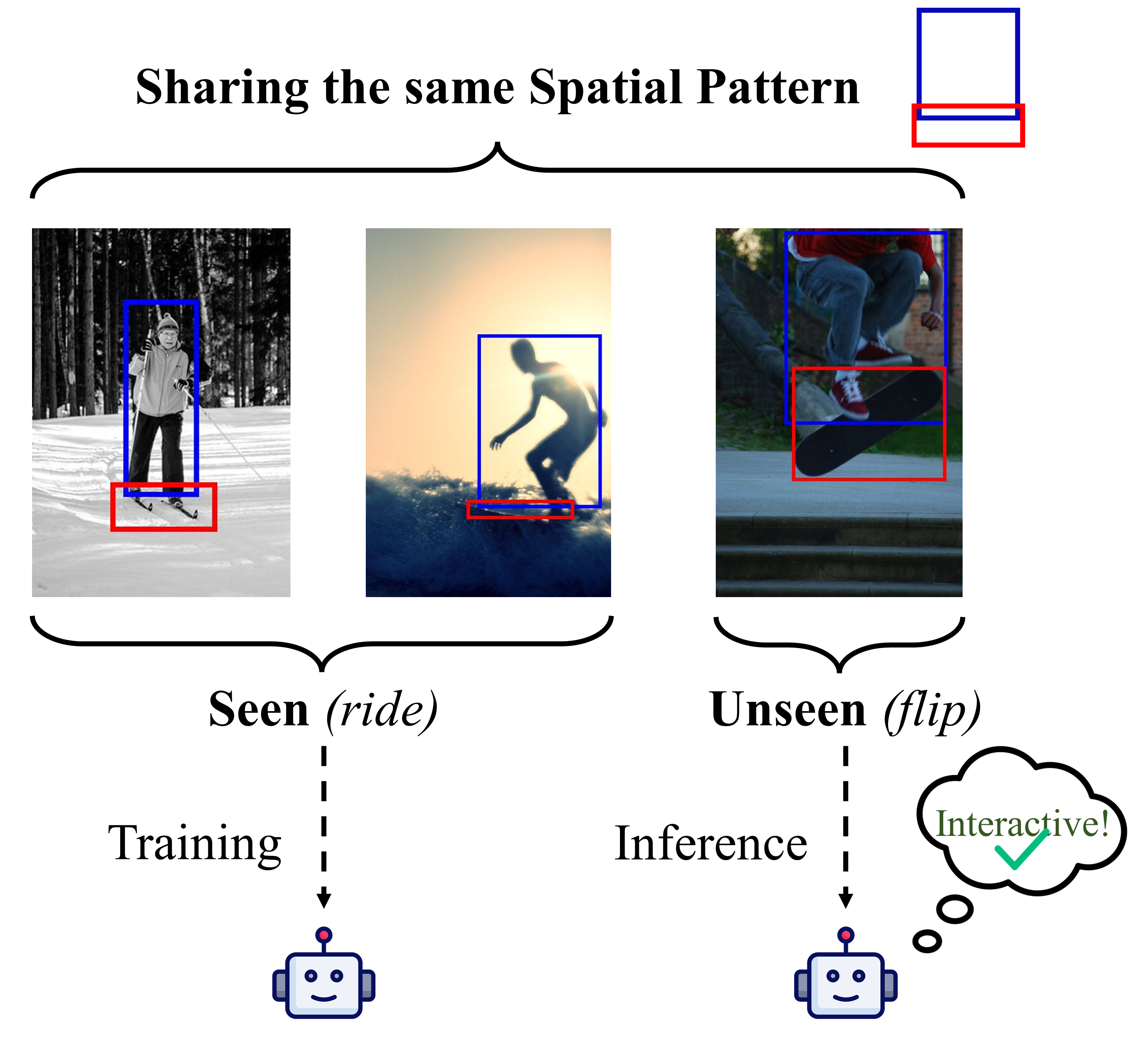}
    \includegraphics[width=0.2\linewidth]{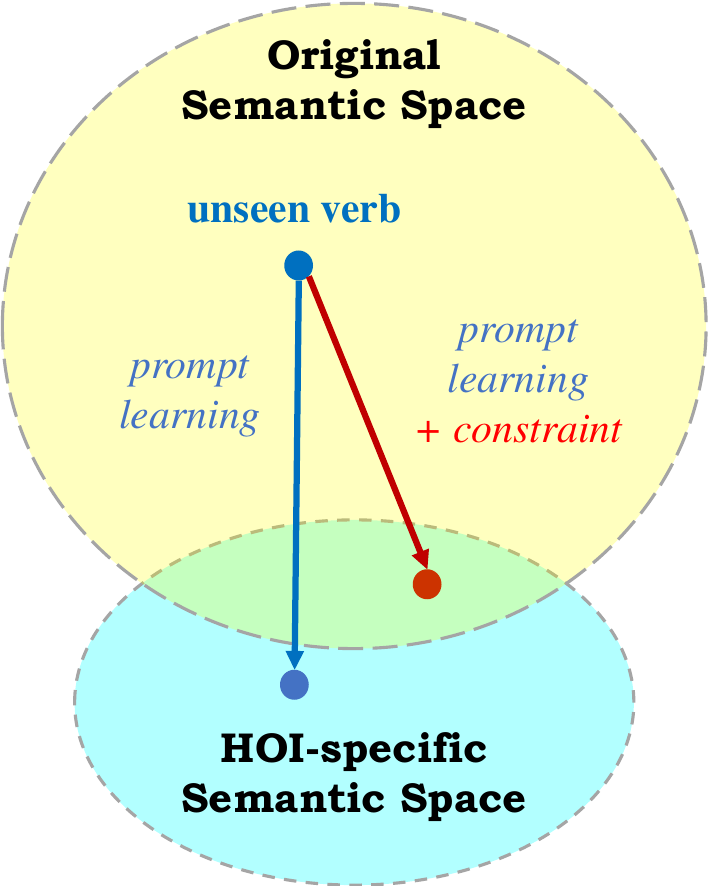}
}
\label{fig:1}
\caption{(a) Previous detectors struggle with the delicate balance between seen and unseen classes, resulting in a low harmonic mean (HM) and poor performance on unseen classes. In contrast, our method effectively addresses this balance issue, leading to significant improvement and establishing a new state-of-the-art benchmark for unseen classes.
(b) Our model uses visual spatial cues during feature extraction to help recognize the interactiveness of unseen HOI concepts and utilize constraint prompt learning for better generalizability on unseen classes.}

\end{figure}

Despite recent advances, most previous works lack generalizability to unseen HOIs. Although some zero-shot HOI detectors~\cite{hou2020visual,hou2021detecting,liao2022gen,wu2022end,wang2022learning} have been proposed in recent years, some of them~\cite{hou2020visual,hou2021detecting} can only detect HOIs of unseen compositions, but fail to incorporate language priors and can't generalize to unseen verbs. 
Besides, many previous zero-shot detectors~\cite{ning2023hoiclip,CLIP4HOI} encounter challenges in achieving a nuanced balance between seen and unseen classes, leading to a low harmonic mean and poor performance on unseen classes, as illustrated in~\cref{fig:degrade_uv}.

Given the combinatorial nature of HOIs, constructing a HOI dataset with all possible HOIs is prohibitively expensive. This motivates us to investigate a HOI detector that can be applied to a wide range of previously unseen interactions with powerful generalizability. Zero-shot HOI detection has the following two challenges: 1) how to extract interactiveness-aware features for human-object pairs to determine whether they interact with each other when confronted with unseen HOI concepts, and 2) how to recognize the unseen interaction types accurately. 

To address the aforementioned issues, \textit{we propose CMMP, which divides HOI detection into two subtasks: interactiveness-aware visual feature extraction and generalizable interaction classification. The design aids in reducing their dependence on one another and error propagation between them.} Inspired by the rapid advancements in large vision-language foundation models and their remarkable zero-shot capability, our objective in this study is to devise a method capable of seamlessly adapting these models to the HOI task using efficient multi-modal prompt learning techniques.

For the first subtask, we introduce conditional vision prompts tailored to guide the extraction of interactiveness- and spatial-aware visual features. This guidance enables the model to generalize its capacity to determine the interactivity of human-object pairs across previously unseen classes.
Specifically, we propose conditional vision prompts incorporating two priors: an input-conditioned instance prior and a global spatial pattern prior. The input-conditioned instance prior encompasses both the spatial configuration and semantics of all detected instances in the input image. This encourages the model to treat both seen and potentially unseen interactive instances equally. As depicted on the left side of~\cref{fig:teasor}, the global spatial pattern prior provides representative plausible spatial configuration of the human and object engaged in interaction, serving as a bridge to discriminate interactivity between seen and unseen HOIs.
Overall, the input-conditioned instance prior offers fine-grained details about individual instances, while the global spatial pattern prior provides a broader contextual understanding of interactions.
As a result, these two conditions contain complementary information that guides spatial-aware visual feature learning for the HOI detection task.
Subsequently, we employ the attention mechanism~\cite{vaswani2017attention} to fuse the above knowledge embedded in conditional vision prompts into the image encoder from the early spatial-aware and fine-grained feature maps, where valuable information lies for the HOI detection task. 
The conditional vision prompts assist the model in extending its capacity to determine the interactivity of human-object pairs from seen categories to unseen ones.

For the subtask of interaction classification, we propose language prompts that are unaware of spatial information. The language prompts provide a unified context for both seen and unseen HOIs, allowing the model to leverage knowledge learned from seen classes to classify HOIs that include unseen verbs. Besides, we use human-designed prompts as a regularizer to keep the learned text prompts from diverging too much. This constraint preserves the origin semantic space learned by large foundation models, as shown on the right of~\cref{fig:teasor}, and thus may be better for potential real-world scenario applications where arbitrary novel actions may occur. 

We propose decoupled vision and language prompts for the above two subtasks to prevent mutual inhibition, respectively. These conditional multi-modal prompts serve as a hub to build a connection between seen and unseen categories. 
We evaluate our detector with conditional multi-modal prompts under various zero-shot settings.Experimental results demonstrate that our method achieves an effective balance between seen and unseen classes, achieving the highest harmonic mean of performance and the best results on unseen classes, as shown in~\cref{fig:degrade_uv}.

Our contributions can be summarized as follows: 
\emph{(1)} To the best of our knowledge, we first propose a multi-modal prompt learning method for large foundation models in zero-shot human-object interaction detection to improve visual-language feature alignment and zero-shot knowledge transfer.
\emph{(2)} Through careful prompt design, we reveal the inherent capacity of the large foundation model for precise discrimination of fine-grained HOIs, enhancing the generalization ability of our CMMP.
\emph{(3)} Our model sets a new state-of-the-art for HOI detection on unseen classes in various zero-shot settings, significantly outperforming all previous methods.

\section{Related Work}
\label{sec:related_work}

\subsection{Human-Object Interaction Detection}

With the development of large-scale datasets~\cite{gupta2015visual,chao2018learning,kuznetsova2020open,liao2020ppdm} and deep learning-based methods~\cite{li2020detailed,ulutan2020vsgnet,tian2023transformer,zhang2024deep,xusemantic,wang2024oed}, HOI learning has been rapidly progressing in two main streams: one- and two-stage approaches.
One-stage HOI detectors~\cite{chen2021reformulating,kim2021hotr,tamura2021qpic,liao2022gen,park2023viplo,kim2023relational,xie2023category,zheng2023open,lei2024exploring} usually formulate HOI detection task as a set prediction problem originating from DETR~\cite{carion2020end} and perform object detection and interaction prediction in a parallel~\cite{chen2021reformulating,kim2021hotr,tamura2021qpic} or sequentially~\cite{liao2022gen}. 
In contrast, two-stage methods~\cite{liu2020amplifying,li2020hoi,zhang2021spatially,zhang2022efficient,zhang2022exploring,zhang2023pvic,ting2023ADA_CM} usually utilize pre-trained detectors~\cite{ren2015faster,he2017mask,carion2020end} to detect human and object proposals and exhaustively enumerate all possible human-object pairs in the first stage. Then they design an independent module to predict the multi-label interactions of each human-object pair in the second stage. 
Despite their improved performance, most previous models rely heavily on full annotations with predefined HOI categories and thus are costly to scale further. Moreover, they lack the generalization capability to deal with unseen HOI categories as shown in~\cref{fig:degrade_uv}. In contrast to them, our work targets zero-shot HOI detection with the help of an off-the-shelf object detector and vision-language model in a two-stage manner.

\subsection{Zero-shot Human-Object Interaction Detection}

Zero-shot HOI detection aims at detecting interactions unseen in the training set, which is essential for developing practical HOI detection systems that can function effectively in real-world scenarios. ConsNet~\cite{liu2020consnet} converts HOI categories and their components into a graph and distributes knowledge among its nodes. VCL~\cite{hou2020visual} recombines object representations and human representations to compose unseen HOI samples. FCL~\cite{hou2021detecting} proposes to generate fake object representations for human-object recombination. 
ATL~\cite{hou2021affordance} exploits additional object datasets for HOI detection to discover novel HOI categories. 
However, lacking the help of semantics, the above methods aren't capable of detecting HOIs including unseen actions. 

To incorporate language priors in zero-shot HOI detection, prevailing approaches~\cite{liao2022gen,wu2022end,ning2023hoiclip,Yuan2022RLIP,CLIP4HOI} propose to incorporate knowledge from CLIP~\cite{CLIP} to achieve zero-shot HOI detection. The natural generalizability of language aids models in recognizing HOIs, even those with unseen actions. 
RLIP~\cite{Yuan2022RLIP} proposes a Relational Language-Image Pre-training strategy for HOI detection. EoID~\cite{wu2022end} distills the distribution of action probability from CLIP to the HOI model and designs an interactive score module combined with a two-stage bipartite matching algorithm to achieve interaction distinguishment. GEN-VLKT~\cite{liao2022gen} utilizes CLIP text embeddings for prompted HOI labels to initialize the classifier and employs CLIP visual features to guide the learning of interactive representations. HOICLIP~\cite{ning2023hoiclip} adopts the one-stage design following GEN-VLKT~\cite{liao2022gen} and proposes query-based knowledge retrieval for efficient knowledge transfer for HOI detection with CLIP. Besides, it exploits zero-shot CLIP knowledge as a training-free enhancement during inference.
Despite the progressing generalizability, previous one-stage methods utilize each query or pair thereof to localize the human-object pair jointly, often leading to overfitting of the decoder to seen categories.

The most relevant work to ours is CLIP4HOI~\cite{CLIP4HOI}, which integrates CLIP into a previously established two-stage method~\cite{zhang2022efficient}, thereby achieving a disentangled two-stage paradigm for zero-shot HOI detection. Different from CLIP4HOI, our method employs conditional multi-modal prompts to directly transform the feature space of large foundation models. This transformation moves from understanding image-level and first-order semantics to comprehending instance-level and second-order semantic information within images, resulting in better generalizability to unseen HOI concepts.

\subsection{Prompt Learning}

Recently, the development of large vision-language models (VLMs), \eg, CLIP~\cite{CLIP}, emerges and finds its applications in various downstream tasks~\cite{zhang2021tip,gao2021clip,zheng2023generating,luo2024zero,yang2024active,liu2023confidence,zheng2024grounding}. Inspired by prompt learning in language tasks, CoOp~\cite{zhou2022learning} first proposes to use context tokens as language prompts in the image classification task. Co-CoOp~\cite{zhou2022conditional} proposes to explicitly condition language prompts on image instances. Recently, other approaches for adapting V-L models through prompting have been proposed. MaPLe~\cite{khattakMaPLe} proposes a coupling function to explicitly condition vision prompts on their language counterparts, to provide more flexibility to align the vision-language representations. However, existing methods primarily focus on prompt learning for image classification, which may not be suitable for HOI detection. 
DetPro~\cite{Du_2022_CVPR} and PromptDet~\cite{feng2022promptdet} propose a novel prompt learning method based on first-order individual instance detection. However, they lack the understanding of second-order pair-wise relationships in images, which is crucial in the HOI detection task.
~\cite{liao2022gen} and~\cite{wang2022learning} first propose applying static template prompts or learnable language prompts in the HOI detection task, respectively. However, they ignore the fact that HOI detection involves considering regional spatial information, making it distinct from image classification.
Therefore, how to design tailored \textit{spatial-aware} prompts specifically designed for the HOI detection task is critical. Given that interactiveness-aware visual feature extraction and interaction classification are distinct subtasks, we propose to employ decoupled multi-modal prompts for these two subtasks to reduce error propagation between them.

\section{Method}

\label{sec:Method}

\subsection{Overview}
\label{subsec:Overview}

HOI detection aims to detect all interactive human-object pairs and predict the interactive relationship for them. Formally, we define the interaction as a quadruple $(b_{h}, b_{o}, a, o)$: $b_{h}, b_{o}$ represent the bounding box of humans and objects and $a \in \mathbb{A}, o \in \mathbb{O}$ represent the human action and object category, where $\mathbb{A} = \{1, 2, ..., A\}$ and  $\mathbb{O} = \{1, 2, ..., O\}$ denote the human action and object set, respectively. Then given an image $\mathbf{I}$, our goal is to predict all quadruples that exist in $\mathbf{I}$. To avoid struggling with multi-task learning~\cite{zhang2021mining} and the risk of overfitting to the joint positional distribution of human-object pairs for seen HOIs~\cite{CLIP4HOI}, we follow the previous two-stage design for HOI detection~\cite{zhang2022efficient,CLIP4HOI}: human-object detection and interaction classification. 
In the first stage, we use an off-the-shelf object detector $\mathrm{D}$, \eg,\ DETR~\cite{carion2020end}, and apply appropriate filtering strategies to extract all instances and exhaustively enumerate the detected instances to compose human-object pairs. 
Then in the second stage, we first encode the image $\mathbf{I}$ using a pretrained image encoder $\mathrm{E_I}$, i.e.,\ $f_{I} = {\mathrm{E_I}}(\mathbf{I}) \in \mathbb{R}^{H\times W\times C}$. We define the union region $b_{u}$ as the smallest rectangular region that contains $b_{h}$, $b_{o}$. Then following the multi-branch architecture of previous HOI detection works~\cite{hou2020visual,hou2021detecting}, we utilize $b_{h}$, $b_{o}$, and $b_{u}$ to extract features for the human region, the object region, and the interaction region from the feature map $f_{I}$ via ROI-align~\cite{he2017mask}, respectively.

\begin{figure}[tb]
    \centering
    \includegraphics[width=0.99\linewidth]{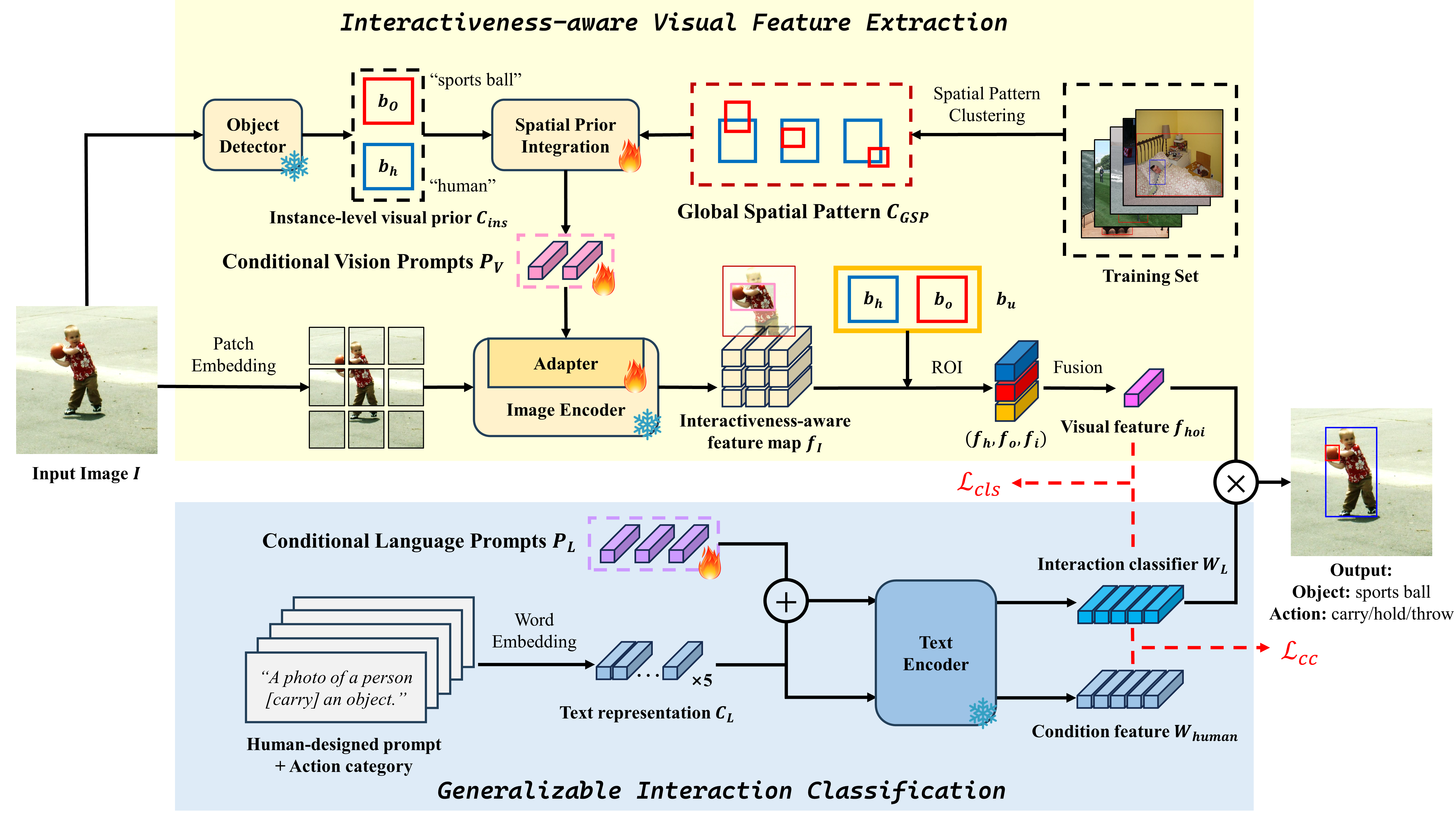}
    \caption{\textbf{The overall framework of CMMP.} The proposed method splits zero-shot HOI detection into two subtasks: interactiveness-aware visual feature extraction and generalizable interaction classification. We propose decoupled vision and text prompts for each subtask to eliminate the dependence between them and break error-propagation in-between. The conditional vision prompts ($P_V$) are used to inject spatial- and interactiveness-aware knowledge into the image encoder and are explicitly constrained by instance-level visual prior ($C_{ins}$) and global spatial pattern ($C_{GSP}$). The conditional language prompts ($P_L$) are constrained by the human-designed prompts ($C_L$) through a regularization loss. (Best viewed in color.)}
    \label{fig:Pipeline}
\end{figure}

As illustrated in~\cref{fig:Pipeline}, CMMP tackles the zero-shot HOI detection task by dividing it into two subtasks: interactiveness-aware feature extraction and interaction classification. To propagate knowledge from seen HOI categories to unseen HOI categories and eliminate the dependence between the two subtasks, we propose decoupled vision prompts $P_{V}$ and language prompts $P_{L}$ for the image encoder $\mathrm{E_I}$ and text encoder $\mathrm{E_{T}}$, respectively.
Specifically, in the image branch, we incorporate instance-level visual prior $C_{ins}$ derived from the input image and global spatial patterns $C_{GSP}$ obtained from the dataset to construct conditional vision prompts $P_{V}$. 
The instance-level visual prior emphasizes the unique characteristics of each detected instance in the input image, encompassing their spatial configurations and semantics. It enables the encoder to treat both seen and potential interactive instances with equal significance. In contrast, the global spatial pattern captures broader relationships and patterns among objects or entities within the scene, creating representative plausible spatial configurations of the human and object under interaction. 
The input-conditioned instance prior furnishes knowledge of individual instances, while the global spatial pattern prior offers a wider contextual comprehension of interactions. Consequently, these two conditions provide complementary guidance for spatial-aware visual feature learning in the HOI detection task.
These prompts are then incorporated into the image encoder to refine its capabilities, transitioning from image-level individual instance comprehension to understanding region-level pair-wise relations. The conditional vision prompts can alert the image encoder $\mathrm{E_I}$ to all potential interactive instances within the image. 
For the text branch, we feed-forward the learnable prompts $P_L$ alongside human-designed prompts $C_L$ to $\mathrm{E_{T}}$. This operation yields the weights of interaction classifier $W_L$, facilitating the computation of interaction scores for the provided human-object pair. Besides, to improve the generalization of large foundation models when fine-tuned on downstream tasks, we incorporate language priors from the pretrained vision-language model by enforcing a consistency constraint between $W_L$ and the condition embeddings $W_{human}$.

\subsection{Interactiveness-aware Visual Feature Extraction}
\label{subsec:interactiveness}

Given that the image encoder of the large foundation model we adopt~\cite{CLIP} is originally trained via Contrastive Language-Image Pre-Training on large-scale image-text pairs, its inherent capability might be limited to grasping image-level first-order semantics.
To endow the image encoder with the capability of distinguishing the interactiveness of all human-object pairs within a given image, we propose to integrate prior knowledge of varying granularity into conditional vision prompts to empower the image encoder to comprehend region-level second-order semantics specifically tailored for the HOI task, as illustrated in~\cref{fig:Pipeline}.

To encourage the model to treat the seen and potentially unseen interactive instances equally, we utilize instance-level information as input-conditioned prior knowledge into the conditional vision prompts. Given an input image $\mathbf{I}$, we first follow ~\cite{ting2023ADA_CM}, utilizing a pretrained object detector to obtain all the instance-level prior knowledge $C_{ins}$ including (1) bounding boxes $b$, which captures the spatial information of detected objects. The spatial configuration might provide cues for understanding interactiveness and is good at transferring as it is instance-agnostic. (2) confidence scores $s$, which reflect the quality and uncertainty of the candidate instances. (3) semantic embeddings $e$ of the detected instances, which are obtained by CLIP language encoder and enable $P_V$ to leverage the category priors to capture which objects can be interacted with. Formally, $C_{ins}$ = MLP(concat($b, s, e$)) $\in$ $\mathbb{R}^{N_{ins}*{d_{ins}}}$, where $N_{ins}$ is the number of detected instances and $d_{ins}$ is the projected feature dimension.

Furthermore, to encourage each instance to be aware of its potential interactive counterpart, we propose integrating global spatial patterns $C_{GSP}$ from the training set with instance-level prior knowledge $C_{ins}$ using a spatial prior integration module $\mathrm{SPI}$. Specifically, for each annotated interactive human-object pair~($b_h^i, b_o^i$), we first compute its unary and pairwise spatial features $sp_i$ used by Zhang et al.~\cite{zhang2021spatially}.
These features include the normalized unary box center, width, and height, as well as pairwise metrics like intersection-over-union, relative area, and direction. More details are provided in the supplementary material. Subsequently, we employ the K-means algorithm to identify the clustering centers of $\{sp_i\}_{i=1}^{N_{hoi}}$ and utilize them as representative spatial patterns of interactive human-object pairs, which are denoted as $C_{GSP}$. 
The global spatial interactive pattern provides a category-agnostic representative plausible spatial configuration between human and object during interaction, 
serving as a bridge to discern the interactivity between seen and unseen HOI concepts. Different from the input-conditioned instance prior that offers fine-grained details about individual instances, the global spatial pattern prior provides a broader contextual understanding of interactions, thus offering supplementary prior knowledge and enhancing the understanding of the interactions.
Overall, the construction of $P_V$ $\in$ $\mathbb{R}^{N_{ins}*{d_{ins}}}$ can be formulated as: 
\begin{equation}
  P_V = \mathrm{SPI}(C_{ins}, C_{GSP}),
  \label{eq:CPV}
\end{equation}
where the spatial prior integration module $\mathrm{SPI}$ is implemented via transformer decoder layers, $C_{ins}$ is treated as query, and $C_{GSP}$ is treated as key and value.
Then, we incorporate $P_V$ into the image encoder $\mathrm{E_I}$ through cross-attention mechanism~\cite{vaswani2017attention}. Formally, we denote $X_{i} \in \mathbb{R}^{hw \times d}$ as the feature map of i-th block of $\mathrm{E_I}$. We first project the feature dimension of $X_i$ to $d_{ins}$ ($d_{ins}$ $<<$ $d$) through $\mathrm{MLP}$:
\begin{equation}
    X_{i}' = {\mathrm{MLP}}(X_{i}),
\end{equation}
where $X_{i}'$ shares the same feature dimension with $P_V$. Then we inject context knowledge $P_V$ into $X_{i}$ through the cross-attention~($\mathrm{Attn}$) mechanism:
\begin{equation}
\label{eq:atten}
    X_i = X_i + {\mathrm{MLP}}({\mathrm{Attn}}(X_{i}', P_V, P_V)),
\end{equation}
where $X_{i}'$ is treated as query and $P_V$ is treated as key and value. The purpose of conditional vision prompts $P_V$ is to utilize its spatial information to enhance the image feature map $X_i$, allowing $X_i$ to acquire valuable spatial information from $P_V$.

\subsection{Generalizable Interaction Classification}
\label{subsec:interaction-classification}
To learn task-specific representations while also retaining generalized CLIP knowledge, we employ language-aware prompt learning with a consistency constraint in the text branch, as shown in~\cref{fig:Pipeline}. The constraint ensures that the learned prototypes of seen and unseen classes leave a reasonable separation margin among each other and do not diverge too far apart.
Specifically, for each action class $a \in \mathbb{A}$, we first format it using the human-designed prompt ``A photo of a person [verb-ing] an object''. We denote $P_L = [P_L^1, P_L^2, ..., P_L^S]$ as the learnable context words, where $S$ denotes the number of learnable prompts. The context words $P_L$ are shared among all classes and thus serve as a bridge between the semantics of seen and unseen categories. The final representation of class $a$ can be obtained by concatenation of learnable context words $P_L$ and the word embedding of the above sentence $C_L^a$.
Then the prototype of the class $a$ can be obtained by the text encoder $\mathrm{E_T}$:
\begin{equation}
    W_L^a = {\mathrm{E_T}}(concat(P_L, C_L^a)), a \in \mathbb{A}
\end{equation}
The prototypes should be the representative features belonging to the corresponding category. Given a sample, the similarity with a prototype could represent how likely it belongs to the category. After performing $l_2$-normalization on all prototypes $W_L^a$, the interaction classifier $W_L$ is then constructed from prototypes of all target classes' embeddings:
\begin{equation}
    W_L = {\mathrm{concat}}(W_L^1, W_L^2, ..., W_L^A)
\end{equation}

To further utilize the feature space learned by the text encoder of VLMs and improve generalization for unseen classes, we propose to use human-designed prompts to guide the feature space of the learnable language prompts. The constraint ensures that prototypes of seen and unseen classes leave a reasonable separation margin among each other and do not diverge too far apart. We apply a regularization loss to reduce the discrepancy between the feature representation of $P_L$ and that of the human-designed language prompts $C_L$. Specifically, we encourage the soft prompt $P_L^i$ to be encoded close to its corresponding human-designed prompt $C_L^i$ through a conditional constraint loss ($\mathcal{L}_{cc}$), which can be formulated as:
\begin{equation}
    \mathcal{L}_{cc} = -\sum_{i=1}^A \log{\frac{\exp(\cos(W_L^i, W_{hum}^i))} 
    {\sum_{j=1}^A \exp(\cos(W_L^i, W_{hum}^j))}},
\end{equation}
where $W_{hum} = {\mathrm{E_T}}(C_L)$ is the encoded features of human-designed prompts $C_L$ and $W_L$ is the feature representation of $P_{L}$.

\subsection{Training CMMP}
Based on the interactiveness-aware feature map $f_{I}$ and the extracted bounding boxes $b_h$, $b_o$, and $b_u$, we first apply ROI-Pooling to extract features for different regions:

\begin{equation}
    f_{hum}, f_{obj}, f_{inter} = {\mathrm{ROI}}(f_{I}, b_h), {\mathrm{ROI}}(f_{I}, b_o), {\mathrm{ROI}}(f_{I}, b_u)
\end{equation}

The interaction classifier $W_L$ is composed of prototypes of all target classes as described in~\cref{subsec:interaction-classification}. We then calculate the action prediction $s_{ho}$ for the corresponding human-object pair as follows: 
\begin{equation}
    s_{ho} = (\alpha_{hum} f_{hum} + \alpha_{obj} f_{obj} + \alpha_{inter} f_{inter}) W_L^T
\end{equation}
where $\alpha_{hum}$, $\alpha_{obj}$, and $\alpha_{inter}$ are set to learnable parameters. We incorporate the object confidence scores into the final scores of each human-object pair. We denote $\sigma$ as the sigmoid function. The final score $s_{ho}^{final}$ is computed as:
\begin{equation}
    s_{ho}^{final} = \sigma(s_{ho}) \cdot (s_h)^\lambda \cdot (s_o)^\lambda,
\end{equation}
where $s_h$ and $s_o$ are confidence scores given by object detector $\mathrm{D}$, and $\lambda > 1$ is a constant that is used to suppress overconfident objects during inference. The whole model is trained on focal loss~\cite{lin2017focal} $\mathcal{L}_{cls}$ for action classification and language regularization loss $\mathcal{L}_{cc}$ at the same time. We use $\lambda_{cc}$ as the hyper-parameter weight. The overall objective function is formulated as:
\begin{equation}
    \mathcal{L} = \mathcal{L}_{cls} + \lambda_{cc} \mathcal{L}_{cc}
\end{equation}

\section{Experiments}

\subsection{Experiment Setting}
\subsubsection{Dataset.}
HICO-DET~\cite{chao2018learning} is a dataset for detecting human-object interactions in images and has 47,776 images (38,118 in train set and 9,658 in test set) and is annotated with $<$\textit{human, verb, object}$>$ triplets. 600 HOI categories in HICO-DET are composed of 80 object classes and 117 verb classes, including no interaction labels.

\subsubsection{Zero-shot Setups.}
To validate our model's zero-shot performance, we evaluate our model on four zero-shot settings on HICO-DET: 1) Unseen Composition (UC), where the training data contains all categories of object and verb but misses some HOI triplet categories. 2) Rare First Unseen Combination (RF-UC)~\cite{hou2021detecting}, which prioritizes rare HOI categories when selecting held-out HOI categories. 3) Non-rare First Unseen Combination (NF-UC)~\cite{hou2021detecting}, which prioritizes non-rare HOI categories instead. Therefore, the training set of the NF-UC setting contains much fewer samples and thus is more challenging. 4) Unseen Verb (UV)~\cite{liao2022gen}, which is set to discover novel categories of actions and reflects a unique characteristic of zero-shot HOI detection.

\subsubsection{Evaluation Metric.}
Following the common evaluation protocol, we use the mean average precision (mAP) to examine the model performance. A detected human-object pair is considered as a true positive if 1) both the predicted human and object boxes have the Interaction-over-Union (IOU) ratio greater than 0.5 with regards to the ground-truth boxes. 2) the predicted HOI categories are accurate.

\subsection{Implementation Details}
We follow the standard protocol of existing zero-shot two-stage HOI detectors~\cite{bansal2020detecting,hou2020visual} to fine-tune DETR on all the instance-level annotations of the training set of HICO-DET before training CMMP.
For the conditional language prompts $P_L$, we set the length of context words $S$ to be 16.
The weight $\lambda_{cc}$ for the consistency loss is set to 1.0 during training. 
We utilize ViT-B/16 as our backbone if not otherwise stated. See more details in the supplementary material.

\subsection{Compare with the State-of-the-art Methods}

We evaluate the performance of our model and compare it with existing zero-shot HOI detectors under UC, RF-UC, NF-UC, UO, and UV settings of the HICO-DET~\cite{chao2018learning} dataset.

As shown in~\cref{tab:ZS}, our CMMP has demonstrated exceptional performance by outperforming all previous detectors by a significant margin on the unseen classes. Furthermore, our CMMP performs comparably to the previous detectors on the seen classes, resulting in an overall outstanding performance.
To be specific, compared to the previous state-of-the-art methods, our CMMP achieves a relative mAP gain of 6.82\%, 3.44\%, 2.07\%, 6.20\%, and 0.81\% on unseen classes on five zero-shot settings, respectively. As shown in the last line of each type in~\cref{tab:ZS}, scaling our CMMP by utilizing the ViT-L/14 backbone to match the FLOPs of CLIP4HOI results in superior performance across all splits.
The performance gap demonstrates our model's ability to excel in both spatial relation extraction for visual features and prototype learning for interaction classification.
Notably, since unseen classes under the NF-UC setting are sometimes more common and semantically straightforward, both our model and previous models~\cite{liao2022gen,wu2022end} may perform better on the unseen split than on the seen split. We also observe that our model performs better on unseen than seen splits in the UO setting, unlike related works. This is likely because CLIP already understands common objects (e.g., bicycles, cars) in the unseen splits. Our consistency constraint preserves CLIP's knowledge, reducing overfitting compared to CLIP4HOI.

Furthermore, previous methods exhibit severe performance degradation between seen and unseen classes, indicating a lack of generalizability. Our model, on the other hand, could alleviate the problem to a large extent and has a high potential for generalization to previously unseen HOIs, confirming the effectiveness of our multi-modal prompts with constraints.

As shown in~\cref{table:fullly-results}, we further compare our CMMP with other methods under the fully supervised setting on HICO-DET and V-COCO datasets. We observe that CMMP improves our baseline model by 5.44 mAP on the full split of HICO-DET, showing the effectiveness of our method design. When scaled to match the FLOPs of CLIP4HOI, our model achieves state-of-the-art performance across all splits of HICO-DET.

\begin{table}[tbh]
    \caption{\textbf{Performance comparison for zero-shot HOI detection.} UC, UO, and UV denote unseen composition, unseen object, and unseen verb settings, respectively. RF- and NF- denote rare first and non-rare first. \#TP/\#AP denotes the number of Trainable/All Parameters. $\dagger$ denotes the scaled-up version utilizing the ViT-L/14 backbone. HM denotes harmonic mean. }

    \label{tab:ZS}
    \centering
    \begin{tabular}{@{}l@{\hspace{4pt}}c@{\hspace{6pt}}c@{\hspace{3pt}}c@{\hspace{3pt}}c@{\hspace{3pt}}c@{\hspace{3pt}}c@{\hspace{3pt}}c@{}}
    \toprule
     Method   & Type & \#TP/\#AP & FLOPs & Unseen$\uparrow$ & Seen$\uparrow$ & Full$\uparrow$ & HM$\uparrow$ \\
     \midrule 
     HOICLIP~\cite{ning2023hoiclip} & UC & - & - & 23.15 & 31.65 & 29.93   & 26.74\\
     CLIP4HOI~\cite{CLIP4HOI}       & UC & 71.2M/262.4M & 186G & 27.71 & \underline{33.25} & \underline{32.11} & 30.23 \\
     \textbf{\textit{CMMP (Ours)}}   & UC & 2.3M/193.4M & 114G & \underline{29.60} & 32.39 & 31.84 & \underline{30.93} \\
     \textbf{\textit{CMMP}$\dagger$ (Ours)} & UC & 5.4M/433.2M & 168G & \textbf{34.46} & \textbf{37.15} & \textbf{36.56} & \textbf{35.75} \\
     \midrule
      GEN-VLKT~\cite{liao2022gen}   & RF-UC & - & - & 21.36 & 32.91 & 30.56   & 25.91\\
      EoID~\cite{wu2022end}         & RF-UC & - & - & 22.04 & 31.39 & 29.52   & 25.90\\
      HOICLIP~\cite{ning2023hoiclip}& RF-UC & - & - & 25.53 & 34.85 & 32.99  & 29.47\\
      CLIP4HOI~\cite{CLIP4HOI}      & RF-UC & 71.2M/262.4M & 186G & 28.47 & \underline{35.48} & \underline{34.08} & \underline{31.59} \\
      \textbf{\textit{CMMP (Ours)}}  & RF-UC & 2.3M/193.4M & 114G & \underline{29.45} & 32.87 & 32.18 & 31.07 \\
      \textbf{\textit{CMMP}$\dagger$ (Ours)} & RF-UC & 5.4M/433.2M & 168G & \textbf{35.98} & \textbf{37.42} & \textbf{37.13} & \textbf{36.69} \\
    \midrule 
      GEN-VLKT~\cite{liao2022gen}      & NF-UC & - & - & 25.05  & 23.38 & 23.71  & 24.19  \\
      EoID~\cite{wu2022end}            & NF-UC & - & - & 26.77  & 26.66 & 26.69  & 26.71  \\
      HOICLIP~\cite{ning2023hoiclip}   & NF-UC & - & - & 26.39  & 28.10  & 27.75  & 27.22 \\
      CLIP4HOI~\cite{CLIP4HOI}         & NF-UC & 71.2M/262.4M & 186G & 31.44 & 28.26 & 28.90  & 29.77 \\
      \textbf{\textit{CMMP (Ours)}}    & NF-UC & 2.3M/193.4M & 114G & \underline{32.09} & \underline{29.71} & \underline{30.18} & \underline{30.85} \\
      \textbf{\textit{CMMP}$\dagger$ (Ours)} & NF-UC & 5.4M/433.2M & 168G & \textbf{33.52} & \textbf{35.53} & \textbf{35.13} & \textbf{34.50} \\
      \midrule
      CLIP4HOI~\cite{CLIP4HOI}       & UO & 71.2M/262.4M & 186G & 31.79 & \underline{32.73} & \underline{32.58} & 32.25\\
      \textbf{\textit{CMMP (Ours)}}  & UO & 2.3M/193.4M  & 114G & \underline{33.76} & 31.15 & 31.59 & \underline{32.40} \\
      \textbf{\textit{CMMP}$\dagger$ (Ours)} & UO & 5.4M/433.2M & 168G & \textbf{39.67} & \textbf{36.15} & \textbf{36.74} & \textbf{37.83} \\
      \midrule
      GEN-VLKT~\cite{liao2022gen}    & UV & - & - & 20.96 & 30.23 & 28.74  &24.76\\
      EoID~\cite{wu2022end}          & UV & - & - & 22.71 & 30.73 & 29.61  &26.12\\  
      HOICLIP~\cite{ning2023hoiclip} & UV & - & - & 24.30 & 32.19 & 31.09  &27.69\\
      CLIP4HOI~\cite{CLIP4HOI}       & UV & 71.2M/262.4M & 186G & 26.02 & 31.14  & 30.42  &28.35\\
      \textbf{\textit{CMMP (Ours)}}  & UV & 2.3M/193.4M  & 114G & \underline{26.23} & \underline{32.75} & \underline{31.84} & \underline{29.13} \\
      \textbf{\textit{CMMP}$\dagger$ (Ours)} & UV & 5.4M/433.2M & 168G & \textbf{30.84} & \textbf{37.28} & \textbf{36.38} & \textbf{33.75} \\
      \bottomrule
    \end{tabular}
\end{table}


\begin{table}[h!]
\caption{\textbf{Performance comparison under fully supervised settings of HICO-DET and V-COCO.} $\dagger$: scaled-up version.}
\label{table:fullly-results}

\centering
\begin{tabular}{@{}l@{\hspace{4pt}}c@{\hspace{4pt}}c@{\hspace{3pt}}c@{\hspace{3pt}}c@{\hspace{3pt}}c@{}}
\toprule
Method & FLOPs & \multicolumn{3}{c@{\hspace{6pt}}}{HICO-DET} & \multicolumn{1}{c@{\hspace{3pt}}}{V-COCO} \\
 & & Rare & Non-rare & Full & $AP^{S2}_{role}$ \\
\midrule
Baseline  & - & 26.64 & 28.15 & 27.80 & 56.2 \\
CLIP4HOI~\cite{CLIP4HOI}  & 186G & \underline{33.95} & \underline{35.74} & \underline{35.33} & \textbf{66.3} \\
\textbf{\textit{CMMP}}    & 114G & 32.26 & 33.53 & 33.24 & 61.2 \\
\textbf{\textit{CMMP}$\dagger$} & 168G & \textbf{37.75} & \textbf{38.25} & \textbf{38.14} & \underline{64.0} \\
\bottomrule
\end{tabular}

\end{table}

\subsection{Ablation Study}

\subsubsection{Network Modules.}
As shown in~\cref{tab:Ablation_module}, we study the effectiveness of different modules of CMMP under the unseen verb setting of HICO-DET. 
We observe the following behaviors related to the use of different modules in the HOID task: \emph{(1)} As shown in lines 1-2 of~\cref{tab:Ablation_module}, in the absence of conditional prompts from one modality, the model's performance is hindered by the inherent limitations of another frozen modality, particularly when dealing with unseen classes. Specifically, the model without conditional language/vision prompts exhibits a 6.83\%, and 5.67\% mAP drop on unseen classes, respectively.
Moreover, the model without conditional vision prompts experiences a clear performance decline in seen classes, compared to the one without conditional language prompts. This is primarily because the image encoder of the foundational model is initially tailored for comprehending image-level, first-order semantics, making it challenging to directly adapt to tasks requiring region-level, second-order relationship understanding.
\emph{(2)} In the absence of any prior knowledge incorporated into the image branch, the model's performance noticeably declines, especially on the unseen classes, as shown in lines 3-4 of~\cref{tab:Ablation_module}. However, when these conditions are combined, our model achieves its optimal performance on both seen and unseen classes, underscoring the complementary nature of these priors in enhancing the generalizability of HOI detection.

\begin{table}[tb]
    \centering
    \begin{minipage}[!t]{0.58\linewidth}
        \caption{\textbf{Ablation on network modules under the Unseen Verb setting.} CLP: Conditional Language Prompts. CVP: Conditional Vision Prompts. GSP: Global Spatial Pattern.
        }
        \label{tab:Ablation_module}
        \begin{tabular}{@{}lccc@{}}
        \toprule
        Setting             & Unseen & Seen & Full \\
        \midrule
        w/o CLP             & 19.40 & 32.13 & 30.35 \\
        w/o CVP             & 20.56 & 26.03 & 25.27 \\
        w/o Instance-level Prior  & 25.07 & 31.27 & 30.40 \\
        w/o GSP             & 24.92 & 31.66 & 30.71 \\
        \textbf{\textit{CMMP (Ours)}}     & \textbf{26.23} & \textbf{32.75} & \textbf{31.84} \\
        \bottomrule
        \end{tabular}
    \end{minipage}\hspace{2.5mm}
    \begin{minipage}[!t]{0.35\linewidth}
        \caption{\textbf{Ablation on the consistency constraint of the language prompts under the Unseen Verb setting.} 
        } 
        \label{tab:Ablation_lamba_CC}
        \centering
        \begin{tabular}{@{}c@{\hspace{6pt}}c@{\hspace{6pt}}c@{\hspace{6pt}}c@{\hspace{6pt}}c@{\hspace{6pt}}c@{}}
        \toprule
        $\lambda_{cc}$ & Unseen & Seen & Full \\
        \midrule
         0     & 24.49 & 32.48 & 31.36 \\
        1.0    & \textbf{26.23} & \textbf{32.75} & \textbf{31.84} \\
        2.0    & 24.45 & 32.24 & 31.15 \\
        \bottomrule
        \end{tabular}
    \end{minipage}
\end{table}

\subsubsection{Constraints for Language Prompts.}
The role of the consistency loss is to serve as a regularization term, allowing the text prompt $P_L$ to learn contextual information through learnable context $U_L$ while preventing excessive deviation from the CLIP text feature space, avoiding a decrease in generalization performance. We conduct experiments using various weights for the consistency loss, as presented in the~\cref{tab:Ablation_lamba_CC}. We observe that: \emph{(1)} When changing the weight $\lambda_{cc}$, the changes in model performance are mainly shown in the unseen categories. This indicates that the regularization loss primarily affects the model's generalization ability. \emph{(2)} When $\lambda_{cc}$ is set to 0, the lack of constraint in the text prompt might cause the textual features to deviate from the CLIP feature space, and decrease the performance on unseen categories. \emph{(3)} As $\lambda_{cc}$ is increased to 1.0, the performance on unseen categories improves, demonstrating an enhancement in model generalization. However, further increasing $\lambda_{cc}$ could potentially result in $U_L$ becoming useless, constraining the model's capacity and leading to a decrease in the final performance. 

\subsection{Qualitative results}
\label{sec:qualitative}
\begin{figure}[htb]
  \centering
  \subfloat[wearing a tie]{
     \label{fig:vis_wear}     
     \includegraphics[width=0.22\linewidth]{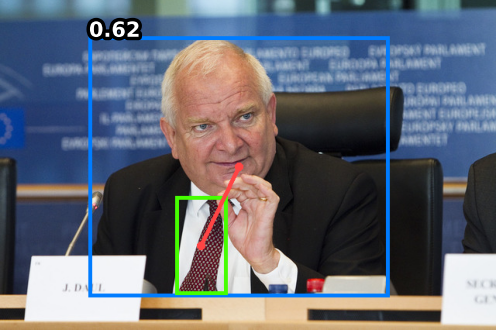}  
    }
  \subfloat[blocking a frisbee]{
     \label{fig:vis_block}     
     \includegraphics[width=0.22\linewidth]{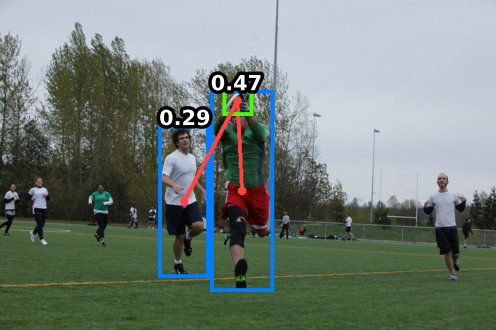}  
    }
  \subfloat[swing a baseball bat]{
     \label{fig:vis_swing}     
     \includegraphics[width=0.22\linewidth]{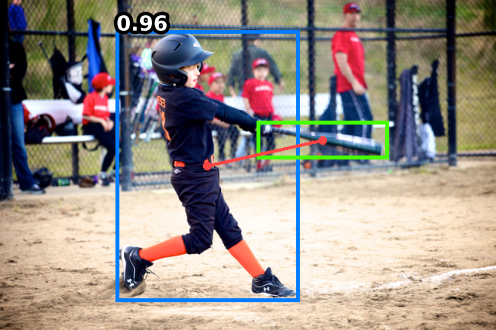}  
    }
  \subfloat[ride a bike]{
     \label{fig:vis_ride}     
     \includegraphics[width=0.22\linewidth]{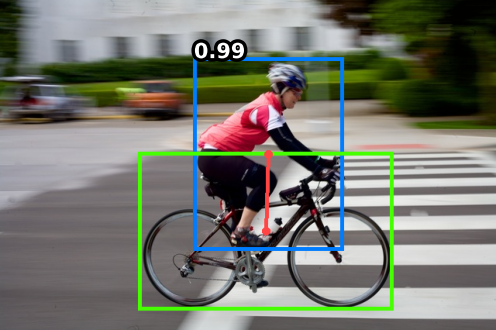}  
    } 
  \caption{Visualization of successfully detected HOIs in the unseen verb setting. Each detected human-object pair is connected by a red line, with the corresponding interaction score overlaid above the human box. All the images contain unseen HOIs made up of unseen verbs and seen objects.}
  \label{fig:vis}
\end{figure}

As shown in~\cref{fig:vis}, we present several qualitative results of successful HOI detections. The visualized HOIs contain unseen verbs, \eg, the verbs ``wear'' and ``swing'' which don't appear in the training set in the unseen verb setting. Our model successfully detects a human-wearing-tie triplet and a human-swing-baseball-bat triplet as shown in~\cref{fig:vis_wear} and~\cref{fig:vis_swing}, which shows the powerful generalizability of our detector.

\section{Conclusion}
We propose CMMP, a novel technique adapting large foundation models for the challenging task of zero-shot HOI detection via conditional multi-modal prompts. 
Our model separates zero-shot HOI detection into two subtasks: extracting spatial-aware visual features and interaction classification, and dealing with them using decoupled multi-modal prompts to break error-propagation in-between.
By carefully designing prompts, we harness the inherent capabilities of large foundation models to precisely discern fine-grained HOIs, thereby enhancing CMMP's generalization ability.
Experimental results across five zero-shot settings show that CMMP outperforms all previous methods by a large margin, establishing a new state-of-the-art for zero-shot HOI detection.

\section*{Acknowledgements}
This work was supported by grants from the National Natural Science Foundation of China (62372014, 61925201, 62132001, U22B2048).

%
%
\bibliographystyle{splncs04}
\bibliography{main}
\end{document}